\documentclass[letterpaper, 10 pt, conference]{ieeeconf}  % Comment this line out if you need a4paper

\IEEEoverridecommandlockouts                              % This command is only needed if 
\usepackage[english]{babel}
\usepackage{amsmath}
\usepackage{amssymb}
\usepackage{booktabs}
\usepackage{siunitx}
\usepackage{multirow}
\usepackage{booktabs}
\usepackage{makecell}
\usepackage{siunitx}
\usepackage{amsfonts}
\usepackage{enumerate}
\usepackage{tabularx}
\usepackage{algorithm,algorithmic}
\usepackage{bm}
\usepackage{adjustbox}
\usepackage{xcolor}
\usepackage{siunitx}
\usepackage{booktabs}
\usepackage{mdframed}
\usepackage{adjustbox}
\usepackage{authblk}
\usepackage{color}
\usepackage{xurl}
\usepackage{subcaption}
\usepackage{cite}
\makeatletter
\let\NAT@parse\undefined
\makeatother
\usepackage{hyperref}
\usepackage{relsize}
\usepackage{float}
\usepackage{pifont}
\DeclareMathOperator*{\argmin}{arg\,min} % thin space, limits underneath in displays

\newcommand{\qbar}{\tfrac{1}{2}\rho V_a^2}
\newcommand{\Tmax}{T_{\max}}

\title{\LARGE \bf Nonlinear Model Predictive Control for Trajectory Tracking of Differentially Flat Fixed-Wing Aerial Systems
}

\author{Nishanth Bobbili$^{*}$, Pratyaksh Rao$^{*}$, Luca Morando, Luca Masci, and Giuseppe Loianno
\thanks{$^*$Equal contribution.}
 \thanks{The authors are with Department of Electrical Engineering and Computer Sciences, University of California, Berkeley, CA 94720, USA.
        {\tt\footnotesize email: \{nishanth.bobbili, pratyaksh10, luca.morando, l.masci, loiannog\}@berkeley.edu}.}
 \thanks{This work was supported by the DARPA Albatross Grant HR00112590173. Approved for Public Release, Distribution Unlimited. The views, opinions and/or findings expressed are those of the authors and should not be interpreted as representing the official views or policies of DARPA or the U.S. Government.}
 }

\begin{document}

\thispagestyle{empty}
\pagestyle{empty}
\makeatletter
\g@addto@macro\@maketitle{
% \vspace{-15pt}
    \setcounter{figure}{0}
    \centering
    \includegraphics[width=\textwidth, trim=0 0 0 0, clip]{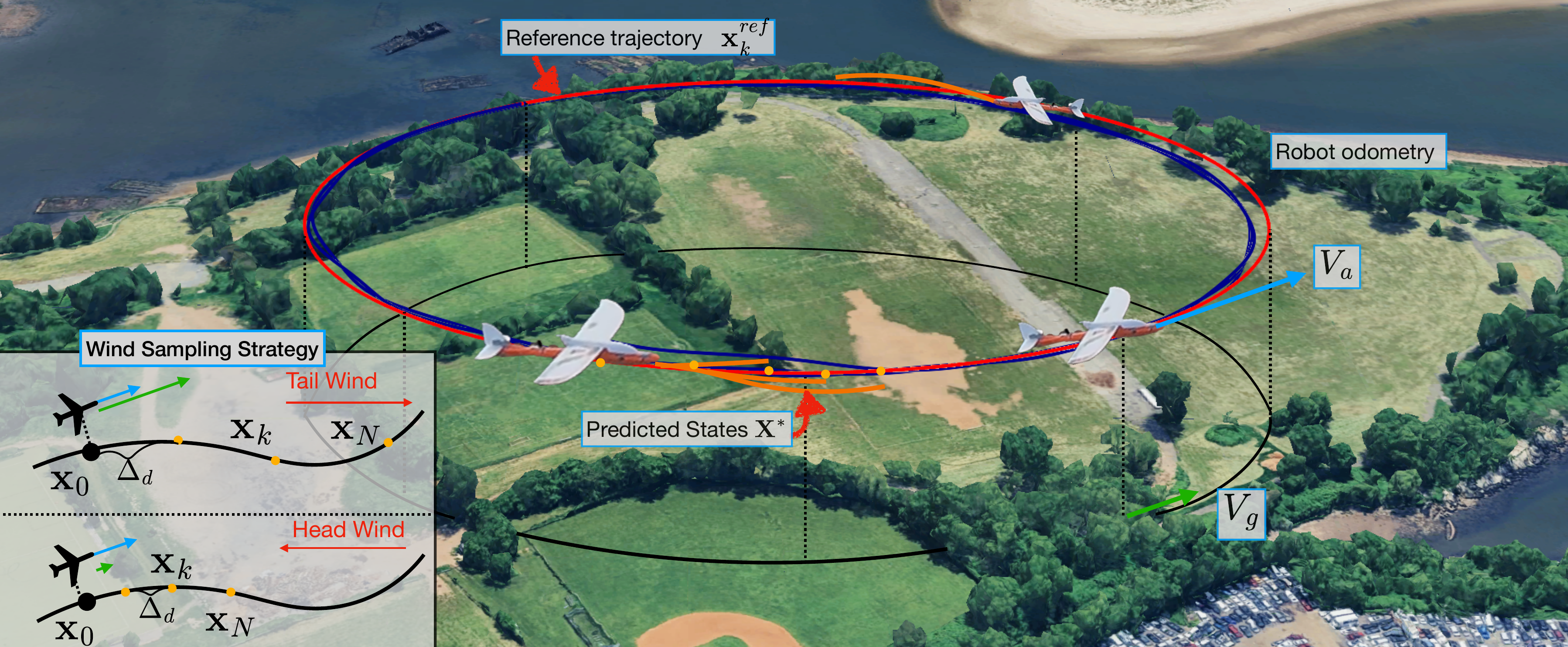}
    \captionof{figure}{Leveraging differential flatness, a wind-aware sampling strategy generates dynamically consistent reference trajectories that adapt to headwind and tailwind conditions. These trajectories are tracked using Nonlinear Model Predictve control, which incorporates aerodynamic forces and moments.\label{fig:contribution}
    %\vspace{-10pt}
    }
}
\makeatother
\maketitle

%%%%%%%%%%%%%%%%%%%%%%%%%%%%%%%%%%%%%%%%%%%%%%%%%%%%%%%%%%%%%%%%%%%%%%%%%%%%%%%%
\begin{abstract}
Planning and control of fixed-wing Unmanned Aerial Vehicles (UAVs) are challenging due to nonlinear dynamics, aerodynamic limits, and environmental disturbances. Differential flatness offers a principled way to generate fast, feasible trajectories, but its use has largely been confined to model-free controllers, which lack predictive capabilities and demand tuning. In this paper, we propose a unified framework that integrates differential flatness-based trajectory generation with Nonlinear Model Predictive Control (NMPC), combining computationally efficient planning with predictive, constraint-aware control. To further improve robustness, we introduce a wind-aware sampling strategy embedded within the NMPC framework, enabling the generation of dynamically feasible reference trajectories that proactively account for wind disturbances while strictly enforcing aerodynamic and control input constraints. We validate the proposed framework through extensive simulations and real-world flight experiments, demonstrating improved tracking accuracy and robustness for complex trajectories, particularly when using the proposed wind-aware sampling strategy under strong wind conditions.

\end{abstract}

%%%%%%%%%%%%%%%%%%%%%%%%%%%%%%%%%%%%%%%%%%%%%%%%%%%%%%%%%%%%%%%%%%%%%%%%%%%%%%%%

\section*{Supplementary material}
\noindent \textbf{Video}: \url{https://youtu.be/Dlz9xllu7RQ?si=BzUoGmHOTVn2YB6o}

\section{Introduction}
Unmanned Aerial Vehicles (UAVs) have demonstrated significant potential across a wide range of applications, including environmental monitoring, disaster response, agriculture, and logistics~\cite{kumar2012opportunities, 10916509}. Among the various designs and configurations of UAVs, conventional Fixed-Wing UAVs (FW-UAVs) remain the preferred platform for many missions due to their superior endurance, extended operational range, and high cruise efficiency~\cite{lyu2023unmanned}.

Despite these advantages, FW-UAVs present unique challenges for planning and control. Their dynamics is highly nonlinear and strongly coupled, with complex aerodynamic effects that are difficult to model accurately. Furthermore, external wind disturbances-often uncertain and time-varying-significantly complicate navigation and trajectory tracking. Together, these factors make planning and control particularly challenging, highlighting the need for reliable, safe, and computationally efficient control frameworks for FW-UAVs.

Traditional flight control architectures decouple lateral and longitudinal dynamics, combining geometric guidance laws~\cite{de2017guidance} with total energy–based control strategies and inner-loop stabilization typically implemented using cascaded PID controllers to drive the aircraft along a prescribed reference path. These  are typically simple geometric primitives (e.g., straight-line segments and circular arcs connected sequentially). While effective and easy to deploy, such frameworks introduce curvature discontinuities, limit maneuver flexibility, and lack predictive capability. Moreover, their performance often degrades under aggressive flight conditions. Classical PID-based implementations are also difficult to tune in highly dynamic regimes and struggle near aerodynamic and actuation limits. In particular, they do not explicitly account for nonlinear aerodynamic effects, actuator saturation, or the strongly coupled dynamics of fixed-wing systems operating near performance boundaries.

\begin{figure*} [t]
    \centering
    \includegraphics[width=\textwidth, trim=0 560 80 0 clip]{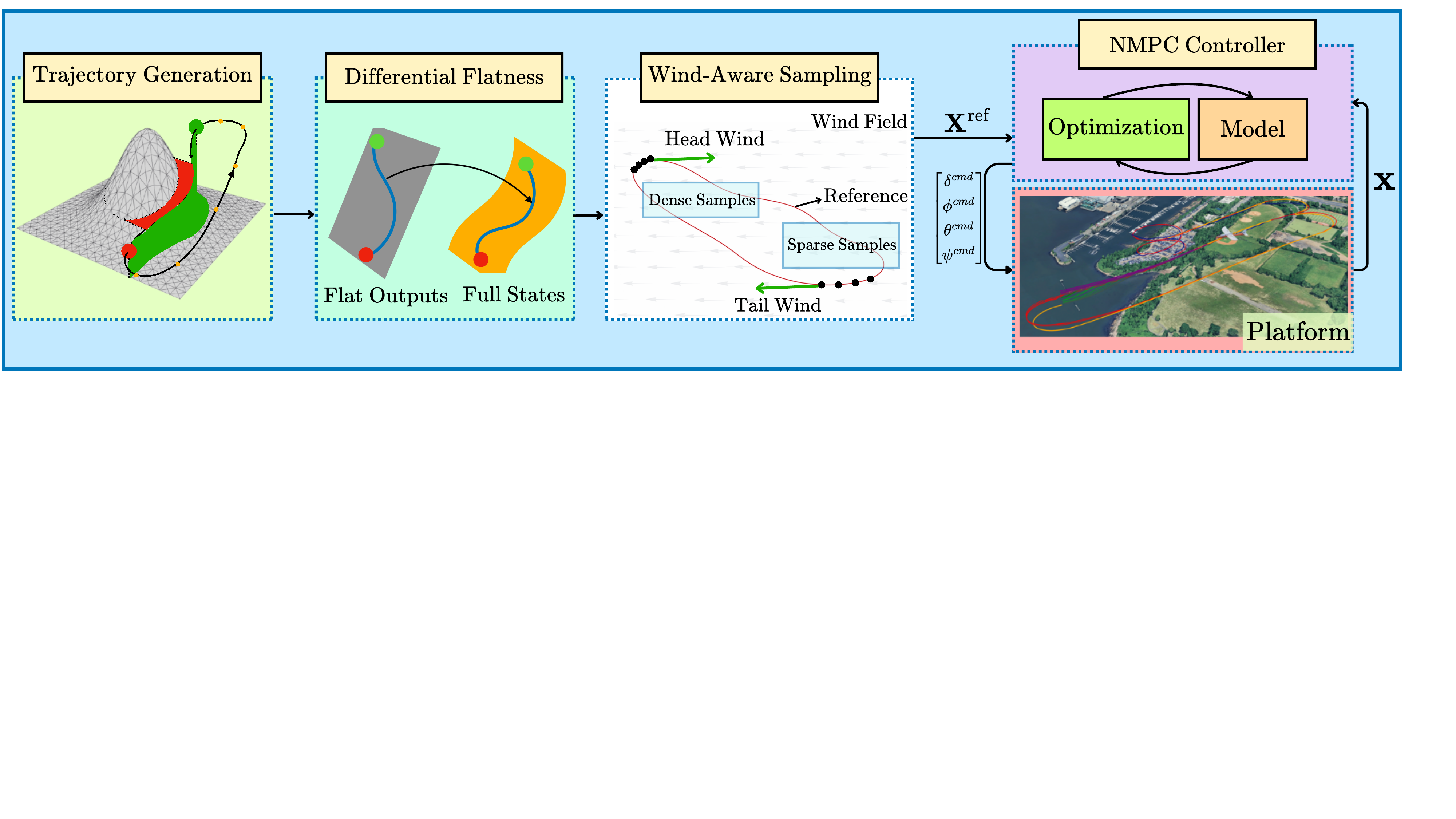}
    \caption{Overview of the proposed NMPC for trajectory tracking.}
    \label{fig:Methodology}
\end{figure*}

These limitations motivate the use of Nonlinear Model Predictive Control (NMPC), which explicitly enforces dynamic feasibility and input constraints while optimizing tracking performance over a receding horizon.
The adoption of NMPC for fixed-wing UAVs has been relatively recent compared to other areas of aerial robotics. This is largely due to the increased modeling complexity associated with strongly coupled nonlinear aerodynamics, sensitivity to wind disturbances, uncertainty in state estimation, limited onboard computational resources, and the practical challenges of experimental validation compared to multirotor platforms. During the past decade, several works have demonstrated the applicability of NMPC to FW-UAVs~\cite{mathisen2015non, stastny2018nonlinear, reinhardt2021nonlinear}, reporting improved robustness and tracking performance compared to classical PID approaches. However, these methods focus on path-following tasks~\cite{drones8020062} or address only lateral or longitudinal control.  Furthermore, generating full-state trajectories can be difficult and also computationally demanding. However, full-state trajectories, rather than purely geometric paths, enable explicit handling of higher-dimensional state constraints and produce dynamically consistent, aggressive references for robotic systems~\cite{Levin,frazzoli2002real}.  A computationally efficient way to generate trajectories is through differential flatness, as extensively demonstrated in the multirotor domain~\cite{5980409,DBLP:journals/corr/abs-2103-00190}. In the fixed-wing setting, the trajectories generated via differential flatness~\cite{morando2025trajectory} satisfy coordinated flight conditions and produce dynamically consistent full-state references with bounded curvature. This opens the possibility of designing an NMPC framework that takes advantage of the flatness property of the system while balancing for the full 6-DOF dynamics. Therefore, we propose an NMPC approach that takes advantage of differentially flatness–derived trajectories to allow curvature-consistent tracking while maintaining robustness under wind disturbances. The proposed approach incorporates the complete vehicle dynamics for high-fidelity trajectory tracking.

Our main contributions can be summarized as follows

\begin{itemize}
    \item We develop an NMPC framework for differentially flat fixed-wing aerial systems, formulated using a full 6-DOF nonlinear model that explicitly incorporates aerodynamic forces, moments, and wind effects at the dynamics level.
    \item We introduce a wind-aware reference generation strategy that adapts the trajectory sampling velocity to guarantee safe cruising airspeed under external disturbances.
    \item We validate the proposed framework through extensive evaluation in simulation and real-world flight experiments, demonstrating improved tracking accuracy and robustness for complex trajectories, particularly when using the proposed wind-aware sampling strategy under strong wind conditions.
\end{itemize}

\section{Related Works}

Navigation of FW-UAVs has been largely relying on guidance controllers combined with cascaded PID loops for path following, where the goal is to steer the vehicle along a desired path or waypoint. The approaches range from geometric methods such as pure pursuit~\cite{yamasaki2009robust} and L1 guidance~\cite{de2017guidance} to control-theoretic methods like Lyapunov-based nonlinear controllers~\cite{lawrence2007lyapunov,chen2009tracking}. These controllers are generally attractive due to their lightweight nature, formal stability guarantees, and ease of deployment on resource-constrained autopilots. However, they cannot naturally account for system dynamics, actuator limits, or sensing constraints, which limits their use in aggressive maneuvers or disturbance-rich environments~\cite{6712082}.

Optimal control techniques such as NMPC have gained significant traction in aerial robotics due to their ability to incorporate actuator and sensor constraints while accounting for the  system dynamics. Applications range from high-level attitude regulation~\cite{6712082, reinhardt2019nonlinear} and low-level actuator control~\cite{reinhardt2023fixed} for FW-UAVs to data-driven and adaptive NMPC formulations~\cite{SavioloTRO2024,RomeroTRO2026} for multirotor platforms. In FW-UAVs, NMPC has been applied to path-following tasks using control-augmented dynamics in conjunction with off-the-shelf autopilots for low-level stabilization~\cite{stastny2018nonlinear}. A general framework for constrained output path-following is presented in~\cite{faulwasser2015nonlinear}, where the system dynamics are augmented with an integrator chain to regulate motion along a parametric curve. More recently, actuator-level NMPC formulations have demonstrated that predictive control can be applied directly at the control surface level while maintaining real-time feasibility~\cite{reinhardt2023fixed}. Despite these advances, the approach is validated primarily in simulation~\cite{reinhardt2023fixed}. Other solutions reduce motion to simplified path primitives and reduced dynamic model~\cite{hirst20253dpathfollowingguidancenonlinear}. In the absence of higher-dimensional full-state references, such strategies remain limited in their ability to fully exploit the flight envelope of fixed-wing platforms. Direct optimization of full-state trajectories~\cite{Levin,frazzoli2002real} can be computationally demanding for onboard implementation.

Trajectory generation in the flat-output space offers a computationally efficient alternative, providing dynamically consistent references without incurring prohibitive optimization costs. Although differential flatness has been leveraged for efficient trajectory generation in FW-UAVs~\cite{morando2025trajectory,bry2015aggressive} typically in combination with PID-based tracking controllers, its integration with fully coupled 6-DOF NMPC formulations has not yet been explicitly addressed. In this work, we unify flatness-based trajectory generation with NMPC in a fully coupled 6-DOF framework. Furthermore, we introduce a wind-aware sampling strategy that anticipates environmental disturbances during reference generation, thereby enhancing robustness under strong wind conditions.

% Frame convention image
\begin{figure} [t]
    \centering
    \includegraphics[width=0.8\linewidth, trim=10 240 640 0, clip]{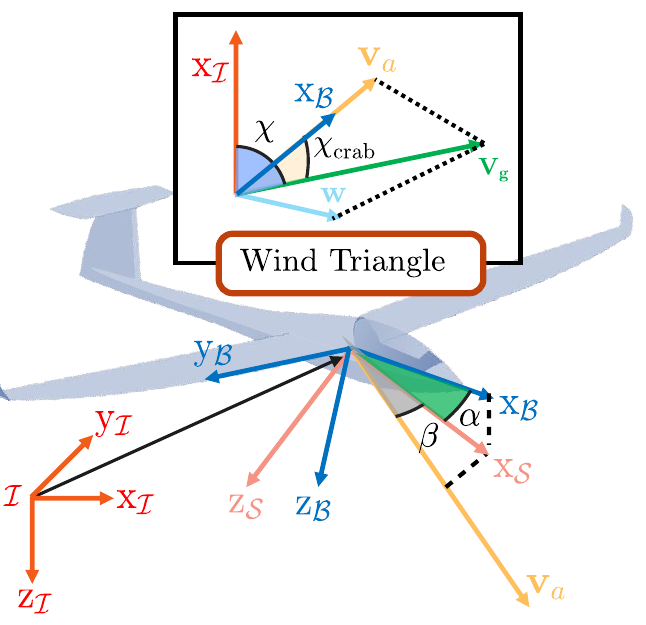}
    \caption{Reference frames and aerodynamic angles.}
    \label{fig:frame_convention}
    %\vspace{-20pt}
\end{figure}

\section{Methodology}
In the following, we present our NMPC formulation as depicted in Fig.~\ref{fig:Methodology}. We first introduce the notation, followed by the system model and the formulation of the associated optimization problem. We then describe the trajectory generation process, including the incorporation of the wind sampling strategy.

\subsection{Preliminaries}
\label{ref:coordinate_frame}

In the following, we consider that bold lowercase symbols denote vectors, like $\mathbf{a}$, while their components are written as non-bold scalars with subscripts, like $a_x, a_y, a_z$. Bold uppercase symbols denote matrices, like $\mathbf{A}$.  Superscript specifies the reference frame in which a vector is expressed, like, $\mathbf{a}^K$ for the $K$-frame and $\mathbf{a}^L$ for the $L$-frame.

We define three coordinate frames for modeling the system, illustrated in Fig.~\ref{fig:frame_convention}. The inertial frame $\{I\}$ is an Earth-fixed North–East–Down (NED) frame with its origin at the home location. The body frame $\{B\}$ is rigidly attached to the UAV. To generate lift, the airfoil of the UAV needs to be oriented at a positive angle with respect to the relative velocity vector $\mathbf{v}_\textbf{a}$, which describes the linear velocity of the UAV with respect to the surrounding air mass, and its corresponding magnitude denotes the airspeed $V_{a}$. This angle is referred to as the angle of attack and denoted by $\alpha \in \mathbb{R}$. The angle of attack is described by a right-handed rotation about $\mathbf{y}^B$ such that the relative velocity vector projected onto the $\mathbf{x}^B-\mathbf{z}^B$ plane is aligned with $\mathbf{x}^B$. We call this resulting coordinate frame the stability frame and is denoted by $\{S\}$. 

\subsection{System Modeling}
 The inertial position is defined as $\mathbf{p}^I = \begin{bmatrix} p_x & p_y & p_z \end{bmatrix}^\top$ and the ground-relative velocity is
$\mathbf{v}^I = \dot{\mathbf{p}}^I = \begin{bmatrix} v_x & v_y & v_z \end{bmatrix}^\top
    \label{eq:inertial_ground_vel}$.
The magnitude of $\mathbf{v}^I$ is the groundspeed $V_g$. The wind vector is denoted as $\mathbf{w}^{I} = [w_x,\, w_y,\, w_z]^\top$. The airspeed $V_a$, is obtained from $\mathbf{v}^I$ as

\begin{equation}
    %V_a = \left\lVert \mathbf{v}^{B} - \mathbf{R}^\top 
    V_a = \left\lVert \mathbf{v}^{I} -  
    \mathbf{w^{I}} \right\rVert_{2}.
    \label{eq:va_definition}
\end{equation}
The attitude $\mathbf{R} \in \mathrm{SO}(3)$ denotes the body-to-inertial frame rotation matrix in the dynamics model, which can be decomposed as
$\mathbf{R} = \begin{bmatrix} \mathbf{r}_x & \mathbf{r}_y & \mathbf{r}_z \end{bmatrix} \in \mathrm{SO}(3),$
where $\mathbf{r}_x, \mathbf{r}_y, \mathbf{r}_z \in \mathbb{R}^3$ are the column vectors of $\mathbf{R}$. This is done by rotating the ground-relative velocity vector expressed in body frame to inertial frame. Accelerations arise from the balance of aerodynamic forces and moments expressed in $\{B\}$. Hence, the rigid-body dynamics are
\begin{equation}
\begin{aligned}
    \dot{\mathbf{p}}^{I} &= \mathbf{R}\,\mathbf{v}^{I}, \\[2pt]
    \dot{\mathbf{v}}^{I} &= \tfrac{1}{m}\mathbf{R}\,\mathbf{f}^{B} 
                          + \mathbf{g}^{I} 
                          , \\[2pt]
    \dot{\mathbf{R}} &= \mathbf{R}\,S\!\left(\boldsymbol{\omega}^{B}\right), \\[2pt]
    \mathbf{J}\,\dot{\boldsymbol{\omega}}^{B} &= S\!\left(\mathbf{J}\,\boldsymbol{\omega}^{B}\right)\boldsymbol{\omega}^{B} + \mathbf{m}^{B},
\end{aligned}
\label{eq:dyn_model}
\end{equation}
where $m$ is the vehicle mass, $\boldsymbol{\omega}^{B} \in \mathbb{R}^3$ is the angular velocity, $\mathbf{f}^{B}\in \mathbb{R}^3$  is the total force, $\mathbf{g}^{I} = [0, 0, g]^\top$ is the gravity vector, $\mathbf{J}\in \mathbb{R}^{3\times 3}$  is the inertia matrix, and $\mathbf{m}^{B}\in \mathbb{R}^3$  are the total aerodynamic moments. The operator $S(\cdot)$ denotes the skew-symmetric matrix associated with the cross product.  
The total force $\mathbf{f}^{B}$ consists of aerodynamic contributions and motor thrust, while the total moment $\mathbf{m}^{B}$ arises from aerodynamic effects about the body axes as
% Aerodynamic forces (with thrust)
\begin{equation}
\mathbf{f}^{B} =
\begin{bmatrix}
\qbar A\!\big[-C_D(\alpha,\omega_y,\delta_e)\cos\alpha 
                \\
\quad + C_L(\alpha,\omega_y,\delta_e)\sin\alpha\big] + \,\delta_t \Tmax \\[6pt]
\qbar A\,C_Y(\beta,\omega_x,\omega_z,\delta_a,\delta_r) \\[6pt]
\qbar A\!\big[-C_D(\alpha,\omega_y,\delta_e)\sin\alpha 
                \\
\quad - C_L(\alpha,\omega_y,\delta_e)\cos\alpha\big]
\end{bmatrix},
\label{eq:forces_body}
\end{equation}

% Aerodynamic moments
\begin{equation}
\mathbf{m}^{B}
=
\qbar A
\begin{bmatrix}
b\,C_l(\beta,\omega_x,\omega_z,\delta_a,\delta_r) \\[6pt]
c\,C_m(\alpha,\omega_y,\delta_e) \\[6pt]
b\,C_n(\beta,\omega_x,\omega_z,\delta_a,\delta_r)
\end{bmatrix},
\label{eq:moments_body}
\end{equation}
where $\rho$, $A$, $b$, and $c$ denote the air density, wing wetted area, wingspan, and mean aerodynamic chord, respectively. The throttle input $\delta_t \in [0,1]$ scales the maximum available motor thrust $\Tmax$. The aerodynamic coefficients $C_L, C_D, C_Y, C_l, C_m, C_n$ are nonlinear functions of the angle of attack $\alpha$, sideslip angle $\beta$, body angular velocity, and control surface deflections $(\delta_a,\delta_e,\delta_r)$. For details on their computation and tabulation, we refer the reader to~\cite{beard2012small}.

\begin{figure} [b]
    \centering
    \includegraphics[width=1.0\linewidth, trim=0 0 0 0, clip]{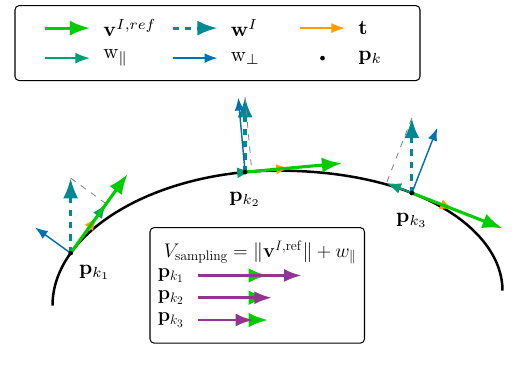}
    \caption{Wind sampling strategy overview.}
    \label{fig:wind_sampling}
\end{figure}

\subsection{Finite-Horizon Optimal Control Problem}
We formulate trajectory tracking as a finite-horizon optimal control problem (OCP). Given $\mathbf{x}(t)$ and $\mathbf{u}(t)$, the current state and control action of the UAV at time $t$, the continuous-time system dynamics are discretized using a fourth-order Runge–Kutta (RK4) scheme with step size $\Delta t$, yielding the discrete-time model

\begin{equation}
    \mathbf{x}_{k+1} = \bar{f}\!\big(\mathbf{x}_{k}, \mathbf{u}_{k}\big)
    = \text{RK4}_{\Delta t}\!\big(f(\mathbf{x}(t), \mathbf{u}(t))\big),
\end{equation}
where $\mathbf{x}_{k}$ and $\mathbf{u}_{k}$ denote the state and input at the discrete time $t_{k} = k \Delta t$. The function, $f(\mathbf{x}(t), \mathbf{u}(t))$ corresponds to the model in eq.~\eqref{eq:dyn_model}. The finite-horizon OCP is then formulated within the direct multiple shooting framework~\cite{bock1984multiple} as
\begin{equation}
\begin{aligned}
    \{\mathbf{X}^*, \mathbf{U}^*\} 
    &= \argmin_{\substack{
        \mathbf{u}_{0}, \ldots, \mathbf{u}_{N-1} \\
        \mathbf{x}_{0}, \ldots, \mathbf{x}_{N}}} 
    \; \sum_{k=0}^{N-1} 
    \!\Big( 
            \|\mathbf{x}_{k} - \mathbf{x}_{k}^{\text{ref}}\|_{Q}^{2} 
            + \|\mathbf{u}_{k} - \mathbf{u}_{k}^{\text{ref}}\|_{R}^{2} 
        \Big)  \\[2pt]
    & \quad + \|\mathbf{x}_{N} - \mathbf{x}_{N}^{\text{ref}}\|_{Q_{N}}^{2} \\[6pt]
    \text{s.t.} \quad 
    & \mathbf{x}_{k+1} = \bar{f}(\mathbf{x}_{k}, \mathbf{u}_{k}), 
    \quad k=0,\ldots,N-1, \\[2pt]
    & \mathbf{x}_{0} = \mathbf{x}(t), \\[2pt]
    & h(\mathbf{x}_{k}, \mathbf{u}_{k}) \leq 0, 
    \quad k=0,\ldots,N-1.
\end{aligned}
\label{eq:nmpc_formulation}
\end{equation}
% where $Q, Q_N \succeq 0$ are weighting matrices on the state deviation, $R \succ 0$ weights control effort, and $\mathbf{x}_k^{\text{ref}}$ denotes the desired reference state.  

\noindent In this formulation, $N \in \mathbb{N}$ denotes the prediction horizon length, i.e., the number of discrete steps over which the optimization is carried out, and $\mathbf{X}^* = [\mathbf{x}_{0}, \ldots, \mathbf{x}_{N}]$ together with $\mathbf{U}^* = [\mathbf{u}_{0}, \ldots, \mathbf{u}_{N-1}]$ represent the optimal state and input sequences. The initial state $\mathbf{x}_{0} = \mathbf{x}(t)$ corresponds to the measured UAV state at the beginning of the horizon. The term $\mathbf{x}_{k}^{\text{ref}}$ denotes the desired reference state at step $k$, and $\mathbf{u}_{k}^{\text{ref}}$ is the nominal control input along the reference trajectory. The weighting matrices $Q$ and $Q_{N}$ penalize state deviations along the horizon and at the terminal step, respectively, while $R$ penalizes deviations in the control inputs. The function $h(\mathbf{x}_{k}, \mathbf{u}_{k}) \leq 0$ encodes state and input inequality constraints, such as actuator limits and flight envelope.

We build the state and control vector using eq.~\eqref{eq:dyn_model} to obtain 
\begin{equation*}
\begin{aligned}
\mathbf{x}_k &= 
\begin{bmatrix}
    {\mathbf{p}_{k}^I}^\top & {\mathbf{v}_{k}^I}^\top & {\mathbf{r}_{x,k}}^\top & {\mathbf{r}_{y,k}}^\top & {\mathbf{r}_{z,k}}^\top  & {\boldsymbol{\omega}_{k}^B}^\top
\end{bmatrix}^\top, \\[6pt]
\mathbf{u}_k &=
\begin{bmatrix}
    \delta_{t,k} & \delta_{e,k} & \delta_{a,k} & \delta_{r,k}
\end{bmatrix}^\top.
\end{aligned}
\end{equation*}
\noindent Additionally, to enforce a user-defined reference airspeed $V^{\text{ref}}_a$, which is the nominal cruise speed of the robot that results in sufficient control authority, we adjust the reference velocity $ {\mathbf{v}_{k}^{I,\text{ref}}}^\top$ based on the wind as explained in the next subsection.

%%%%%%%%%%%%%%%%%%%%%%%%%%%%%% FIGURE: Simulation DYNAMICS %%%%%%%%%%%%%%%%%%%%%%%%%%%%%%%%%%%%%%%%%%%%
\begin{figure*}[t]
\vspace{-0.2cm} 
  \centering
  % First panel
  \begin{minipage}[t]{0.5\textwidth}\centering
    \includegraphics[height=6.5cm,trim=0 0 00 0,clip]{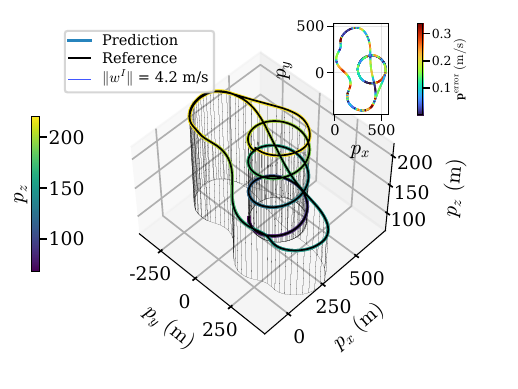}
    \subcaption{Position tracking: prediction vs. reference.}
  \end{minipage}\hfill
  % Second panel
  \begin{minipage}[t]{0.5\textwidth}\centering
    \includegraphics[height=6.5cm,trim=0 0 0 0,clip]{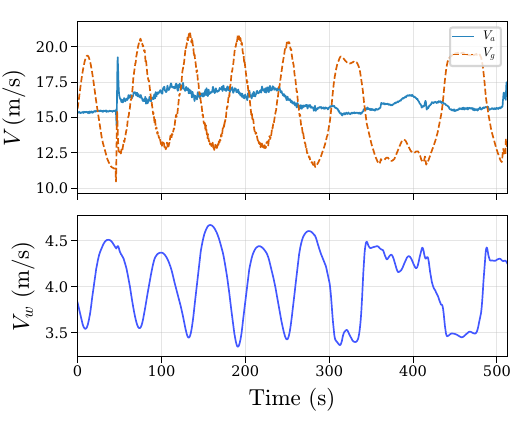}
    \subcaption{Airspeed vs. groundspeed with wind-state estimation.}
  \end{minipage}
  \caption{NMPC tracking results with wind up to $4~\text{m/s}$.}
  \label{fig:nmpc_results_pos_wind}
\end{figure*}

%%%%%%%%%%%%%%%%%%%%%%%%%%%%%% FIGURE: Simulation DYNAMICS %%%%%%%%%%%%%%%%%%%%%%%%%%%%%%%%%%%%%%%%%%%%
\subsection{Trajectory Generation and Wind-aware Sampling}
%We leverage differential flatness under coordinated flight to 
Reference trajectories are constructed using a differential flatness-based planning framework and are parameterized with respect to the path length~\cite{morando2025trajectory}. A dynamical system is considered differentially flat if its states and control inputs can be expressed as functions of a set of flat outputs and a finite number of their derivatives.
For the fixed-wing platform described by the coordinated flight model, the dynamics can be transformed into a feedback-linearizable form~\cite{hauser1997aggressive}, where the flat outputs are represented by the position components of $\mathbf{p}^{I}$. We generate smooth state/reference trajectories in output space. These serve as references (and warm starts) for an NMPC that enforces feasibility under the full model and actuator constraints. The flat outputs are mapped through dynamic inversion to reconstruct the full reference state. Trajectories in the flat output space are generated through a convex Quadratic Programming (QP) problem using Bernstein polynomials.

The reference ground-relative velocity ${\mathbf{v}^{I,\text{ref}}}^\top$ from the trajectory is defined under a no-wind condition and therefore must be adjusted to account for the wind. To achieve this, we decompose the vector ${\mathbf{w}^I}^\top$ into parallel and perpendicular components, $\text{w}_\parallel$ and $\text{w}_\perp$ respectively with respect to the unit direction vector $\mathbf{t} = \mathbf{v}^{I,\text{ref}} / \|\mathbf{v}^{I,\text{ref}}\|$ as shown in Fig.~\ref{fig:wind_sampling}. 
\begin{equation}
\text{w}_{\parallel} =
{\mathbf{w}^I}^\top \mathbf{t},
\label{eq:wind_parallel}
\end{equation}

\begin{equation}
    \text{w}_{\perp} =
    \mathbf{w}^I - \text{w}_\parallel \mathbf{t}. \label{eq:wind_perpendicular}
\end{equation}
We then add the parallel component to the reference velocity to determine the progress velocity used to sample the reference horizon 
\begin{equation}
    V_{\text{sampling}} = ||\mathbf{v}^{I,\text{ref}}|| + \text{w}_{\parallel}.
\end{equation}
The wind vector $\mathbf{w}^I$ is assumed constant over the prediction horizon, since the optimizer runs at $100$~Hz, a timescale substantially faster than that of wind variations, which particularly at low altitudes, typically evolve over several seconds or more. We also assume that $||\mathbf{w}^I||$ is never greater than $\mathbf{v}^{I,\text{ref}}$ to ensure that the robot always maintains a minimum positive groundspeed. The effect of this adjustment is depicted in Fig.~\eqref{fig:Methodology} showing the length of $\mathbf{v}^{I,\text{ref}}$ lengthening in the tail-wind and shortening in the headwind. Since the trajectories are parameterized by distance, they can therefore be sampled at discrete intervals $\Delta d = V_{\text{sampling}} \, \Delta t$.

%%%%%%%%%%%%%%%%%%%%%%%%% Constraints %%%%%%%%%%%%%%%%%%%%%%%%%%%%%%%%%%%%%%%

\begin{table}[t]
\centering
\caption{NMPC parameters and constraints.}
\label{tab:nmpc_summary}
\renewcommand{\arraystretch}{1.15}
%\setlength{\tabcolsep}{5pt}

% ===================== CONSTRAINTS =====================
\begin{tabular}{l c}
\toprule
\textbf{Constraint} & \textbf{Bounds} \\
\midrule
$(p_x, p_y, p_z)$ [m] & $[-10^{4},\,10^{4}]$ \\
$(v_x, v_y, v_z)$ [m/s] & $[5,\,25]$ \\
$(\omega_x, \omega_y, \omega_z)$ [rad/s] & $[-10,\,10]$ \\
$\delta_a$ [deg] & $[-30,\,30]$ \\
$\delta_e$ [deg] & $[-30,\,30]$ \\
$\delta_r$ [deg] & $[-30,\,30]$ \\
$\delta_t$ & $[0,\,1]$ \\
Horizon length $N$ & $10$ \\
Horizon time $T$ [s] & $2$ \\
\bottomrule
\end{tabular}
\end{table}

%%%%%%%%%%%%%%%%%%%%%%%%% Constraints %%%%%%%%%%%%%%%%%%%%%%%%%%%%%%%%%%%%%%%

%\begin{table}[t]
%\centering
%\caption{Position RMSE (mean and variance) under varying wind magnitudes, averaged over 5 trials.}
%\label{tab:rmse_wind}
%\renewcommand{\arraystretch}{1.15}
%\setlength{\tabcolsep}{5pt}
%\begin{tabular}{c|cc|cc}
%\toprule
%\multirow{2}{*}{Wind [m/s]} 
%& \multicolumn{2}{c|}{Kinematic NMPC} 
%& \multicolumn{2}{c}{Dynamics NMPC} \\ 
%& Mean [m] & Var [m] & Mean [m] & Var [m] \\
%\midrule
%$2$  & $1.736$   & $0.006$  & $1.21$ & $0.08$ \\
%$4$  & $1.749$   & $0.020$   & $1.76$ & $0.15$ \\
%$6$  & $1.917$   & $0.111$   & $1.96$ & $0.18$ \\
%$8$  & $1.923$   & $0.035$   & $2.36$ & $0.21$ \\
%$10$ & $2.529$   & $0.475$  & $3.49$ & $0.26$ \\
%\bottomrule
%\end{tabular}
%    \vspace{-10pt}
%\end{table}

%\vspace{-3pt}
\section{Experimental Setup}

%\subsection{System setup}
Real-world experiments are conducted using a custom Strix Stratosurfer UAV equipped with an onboard computer, the NVIDIA Orin NX (Eight-core (ONX 16GB) Cortex A78AE Armv8.2(64-bit)) running Ubuntu 22.04 and the ROS\footnote{\url{www.ros.org}} framework for intra-processes communications. The computer is connected via UART to an STM32H757-based flight controller (Orange+), manufactured by CubePilot and running PX4 Autopilot. The UAV employs an airspeed sensor and a GNSS module (a Drotek\textsuperscript{{\textregistered}} F9P GNSS), fused through the PX4 EKF2 estimator for state estimation and localization.  Along with the UAV state, the flight controller also provides estimates of the airspeed $V_{a}$ and the wind velocity in the x-y plane of the inertial frame (i.e., $\text{w}^I_x$ and $\text{w}^I_y$). 

\begin{figure*}[t]
  \centering
  \begin{minipage}[t]{0.5\textwidth}
   \centering
\includegraphics[width=\linewidth,keepaspectratio,trim=0 0 0 0, clip]{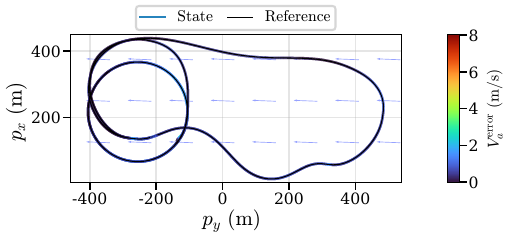}
    \subcaption{NMPC with Wind-aware Sampling.}
  \end{minipage}\hfill
  \begin{minipage}[t]{0.5\textwidth}
    \centering
    \includegraphics[width=1\columnwidth,keepaspectratio,trim=0 0 0 0, clip]{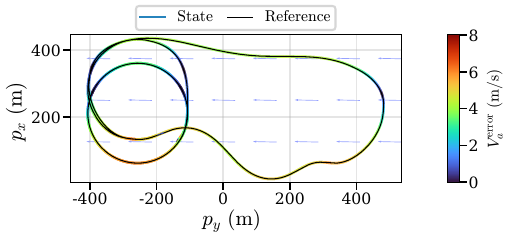}
    %\hspace*{-1.0cm}
    \subcaption{NMPC without Wind-aware Sampling.}
  \end{minipage}
  \caption{Airspeed tracking performance with and without wind-aware sampling in 7m/s wind.}
  \label{fig:with_vs_without_sampling}
\end{figure*}

We leverage \texttt{acados}~\cite{verschueren2022acados} to generate efficient C code for real-time onboard implementation with $N=10$ shooting steps, covering the evolution of the system over $2~\si{\second}$. By utilizing dynamically feasible full-state references, we significantly improve optimizer convergence and solution quality.
Rather than adopting the Real-Time Iteration (RTI) scheme of Sequential Quadratic Programming (SQP), we choose to solve the nonlinear program to convergence at each control step, thereby preserving optimality. Despite this choice, the controller operates at frequencies of around 100\,Hz onboard the vehicle. The resulting quadratic subproblems are solved using the high-performance interior-point method (HPIPM)~\cite{frison2020hpipm}. The constraints imposed in the optimization problem are summarized in Table~\ref{tab:nmpc_summary}.
The aerodynamic coefficients are identified using Athena Vortex Lattice (AVL)~\cite{budziak2015aerodynamic}, where the UAV geometry is modeled to extract aerodynamic parameters. The reference control inputs are set to the aircraft trim condition corresponding to steady, level cruise flight: throttle $\delta_t = 0.60$, elevator $\delta_e = 8.0^{\circ}$, aileron $\delta_a = 0.0^{\circ}$, and rudder $\delta_r = 0.0^{\circ}$. These trim values define the nominal equilibrium about which the NMPC optimizes actuator deflections. 
Although the optimization is formulated and solved at the individual control input level, as presented in eq.~(\ref{eq:dyn_model}), the NMPC controller transmits collective thrust and angular rate setpoints derived from the optimal solution. The PX4 low-level flight controller then executes the inner-loop attitude and thrust regulation through its onboard PID stabilization framework. This design choice is motivated by practical limitations of low-level flight controllers, which restrict effective actuator command update frequencies to below approximately 50~Hz\footnote{\url{https://docs.px4.io/main/en/}} . As a result, directly transmitting high-frequency motor-level commands is impractical and can reduce responsiveness. Instead, providing higher-level control setpoints enables faster and more stable control execution. Moreover, model mismatches in real-world systems may induce actuator chatter when individual motor inputs are commanded directly, further motivating the use of thrust and body-rate setpoints. Both simulations and real-world experiments employ the same dynamic model, solver, and model parameters.

\vspace{-5.5pt}

%Hence, we transmit the thrust and angular rate setpoints derived from the NMPC optimal states, while still solving the optimization problem over the full dynamic model.

% First figure in left column
% \begin{figure}[t]
%     \centering
%     \includegraphics[width=0.9\linewidth, trim=0 0 0 0, clip]{images/results/simulation/wind_ablation.pdf}
%     \caption{Position tracking RMSE under varying wind magnitudes in SITL simulation. The solid line shows the mean RMSE over 5 trials, with the shaded region indicating the corresponding $\pm 1\sigma$ variation.}
%     \label{fig:wind_ablation}
%     \vspace{-8pt}
% \end{figure}

\section{Results}
In this section, we demonstrate the feasibility of solving the trajectory tracking problem for FW-UAVs.
We evaluate the formulation in two scenarios: (a) software-in-the-loop (SITL) simulations using PX4 with Gazebo as the physics engine, and (b) real-world flight experiments. 

%The experiments are designed to address two central questions - a) Are both NMPC formulations capable of achieving long-horizon, closed-loop trajectory tracking on fixed-wing UAVs in realistic flight conditions? and b) How robust are the NMPC formulations to external disturbances such as wind?

% The experiments were conducted to evaluate the robustness of the controller within a test area of $0.15~\text{km}^2$, under wind conditions ranging from $3$ to $10~\text{m}\,\text{s}^{-1}$. The UAV is commanded to shuttle between $2$ and $5$ waypoints, performing circular orbits of up to $150$\,m radius. The waypoints are distributed at different altitudes, enabling assessment of the controller’s climb and descent performance.
 %When the climb rate exceeded a specified threshold, the reference trajectory was modified to a spiral pattern. %The experiments were conducted under wind conditions with magnitudes of up to $8$ \,m/s

\subsection{Robustness to Wind in SITL Simulation}
% We first investigate how effectively both formulations handle external disturbances in simulation. Since wind cannot be systematically controlled in real-world experiments, we conduct this test in SITL by varying the wind magnitude directly in the simulator. Table~\ref{tab:rmse_wind} presents an ablation of the position RMSE as a function of wind strength. For each wind setting, we run $5$ independent trials and report the mean and standard deviation ($±1\sigma$) spread. As expected, the RMSE increases with stronger winds due to the higher disturbance force. However, in all cases, the controllers remain stable, with the overall RMSE remains less than $3.19$~m m even at the highest tested wind levels of $10~\text{m}\text{s}^{-1}$, highlighting the robustness of both formulations in maintaining accurate trajectory tracking under significant disturbances. 

Fig.~\ref{fig:nmpc_results_pos_wind} illustrates spiral trajectory tracking using NMPC formulation under wind with wind-aware sampling. The 3D plots show the reference trajectory and the NMPC-predicted trajectory, with the inset highlighting the top-down x-y view. The plotted case corresponds to a spiral maneuver, which requires coordinated changes in roll, pitch, and velocity. Despite the complexity of the maneuver, state remains close to the reference, demonstrating that the controller preserve trajectory fidelity under sustained wind. 

\begin{figure} [t]
    \centering
    \includegraphics[width=1\linewidth, trim=0 0 0 0, clip]{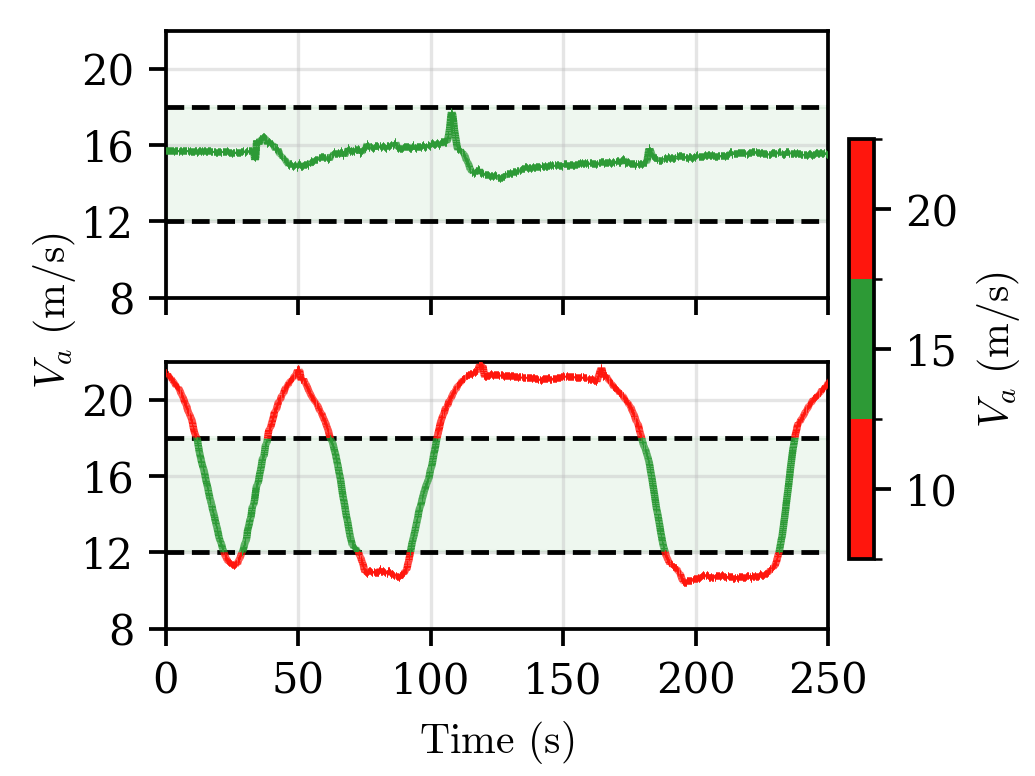}
    \caption{Airspeed tracking with wind adjustment (top) vs Airspeed tracking without wind adjustment (bottom).}
    \label{fig:wind adjusted airspeed tracking}
\end{figure}
\begin{table}[h]
\centering
\caption{NMPC airspeed tracking and AoA performance}
\label{tab:nmpc_airspeed_aoa}
\setlength{\tabcolsep}{5pt}
\renewcommand{\arraystretch}{1.15}
\footnotesize
\begin{tabular}{c c c c c}
\toprule
&
\multicolumn{2}{c}{\textbf{NMPC with sampling}} &
\multicolumn{2}{c}{\textbf{NMPC w/o sampling}} \\
\cmidrule(lr){2-3}\cmidrule(lr){4-5}
\makecell{$\|\mathbf{w}^I\|$\\$[\si{m/s}]$} &
\makecell{Mean $V_a$\\error $[\si{m/s}]$} &
\makecell{Max AoA\\$[\si{\degree}]$} &
\makecell{Mean $V_a$\\error $[\si{m/s}]$} &
\makecell{Max AoA\\$[\si{\degree}]$} \\
\midrule
$2$  & $0.874$ & $4.949$ & $1.274$ & $5.490$ \\
$4$  & $0.867$ & $6.383$ & $2.317$ & $7.685$ \\
$6$  & $0.541$ & $6.839$ & $3.408$ & $11.016$ \\
$8$  & $0.379$ & $8.180$ & $4.514$ & $15.514$ \\
$10$ & $0.592$ & $9.389$ & $5.404$ & $20.967$ \\
\bottomrule
\end{tabular}
\end{table}
%  Infeasible Trajectory
\begin{figure} [t]
    \centering
    \includegraphics[width=1\linewidth, trim=0 0 0 0, clip]{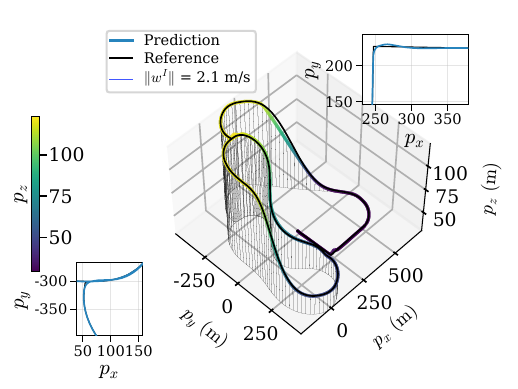}
    \caption{Recovery from unfeasible reference trajectory.}
    \label{fig:Infeasible Rerference}
\end{figure}

We compare the tracking performance of the NMPC with and without the proposed wind-aware sampling strategy in Table~\ref{tab:nmpc_airspeed_aoa}. The results show that the NMPC with wind-aware sampling achieves significantly better airspeed tracking than the baseline NMPC without sampling. In particular, the proposed approach tracks the desired airspeed with an average error of approximately 1 m/s, as shown in Fig.~\ref{fig:with_vs_without_sampling}.

Table~\ref{tab:nmpc_airspeed_aoa} also reports the mean airspeed error and the achieved and maximum angles of attack (AoA) for both NMPC variants under identical wind conditions. The results indicate that, without the proposed sampling strategy, the NMPC allows substantially larger AoA deviations, bringing the aircraft close to stall. Furthermore, Fig.~\ref{fig:wind adjusted airspeed tracking} shows that the controller without wind-aware sampling can drive the aircraft into unsafe operating airspeeds, whereas the proposed method maintains safer and more consistent tracking behavior.

% \begin{table}[h]
% \centering
% \caption{NMPC airspeed tracking performance}
% \label{tab:nmpc_mean_std}

% \setlength{\tabcolsep}{4pt}
% \renewcommand{\arraystretch}{1.15}
% \footnotesize

% \begin{tabular}{c c c c c}
% \toprule
% &
% \multicolumn{2}{c}{\textbf{NMPC with sampling}} &
% \multicolumn{2}{c}{\textbf{NMPC w/o sampling}} \\
% \cmidrule(lr){2-3}\cmidrule(lr){4-5}
% \makecell{$\|\mathbf{w}^I\|$\\$[\si{m/s}]$} &
% \makecell{Mean $V_a$ error\\$[\si{m/s}]$} &
% \makecell{Max position\\error $[\si{m}]$} &
% \makecell{Mean $V_a$ error\\$[\si{m/s}]$} &
% \makecell{Max position\\error $[\si{m}]$} \\
% \midrule
% $2$  & $0.874$ & $1.237$ & $1.274$ & $1.750$ \\
% $4$  & $0.867$ & $1.940$ & $2.317$ & $1.680$ \\
% $6$  & $0.541$ & $1.780$ & $3.408$ & $3.060$ \\
% $8$  & $0.379$ & $1.820$ & $4.514$ & $2.900$ \\
% $10$ & $0.592$ & $1.660$ & $5.404$ & $5.028$ \\
% \bottomrule
% \end{tabular}

% \vspace{-10pt}
% \end{table}

%Although, it can be seen in Fig.~\ref{fig:kinematic_nmpc_results_pos_thrust} that the quality of the predictions degrade with poor wind state estimation from the PX4. Nevertheless, despite the increase in error, both formulations are able to track the reference trajectory successfully even under high wind conditions.
%%%%%%%%%%%%%%%%%%%%%%%%%%%%%% FIGURE: REAL WORLD DYNAMICS %%%%%%%%%%%%%%%%%%%%%%%%%%%%%%%%%%%%%%%%%%%%

\begin{figure*}[t]
  \centering
  \begin{minipage}[t]{0.5\textwidth}
    \centering
    \includegraphics[height=8cm,width=\linewidth,keepaspectratio,trim=5 10 1 10, clip]{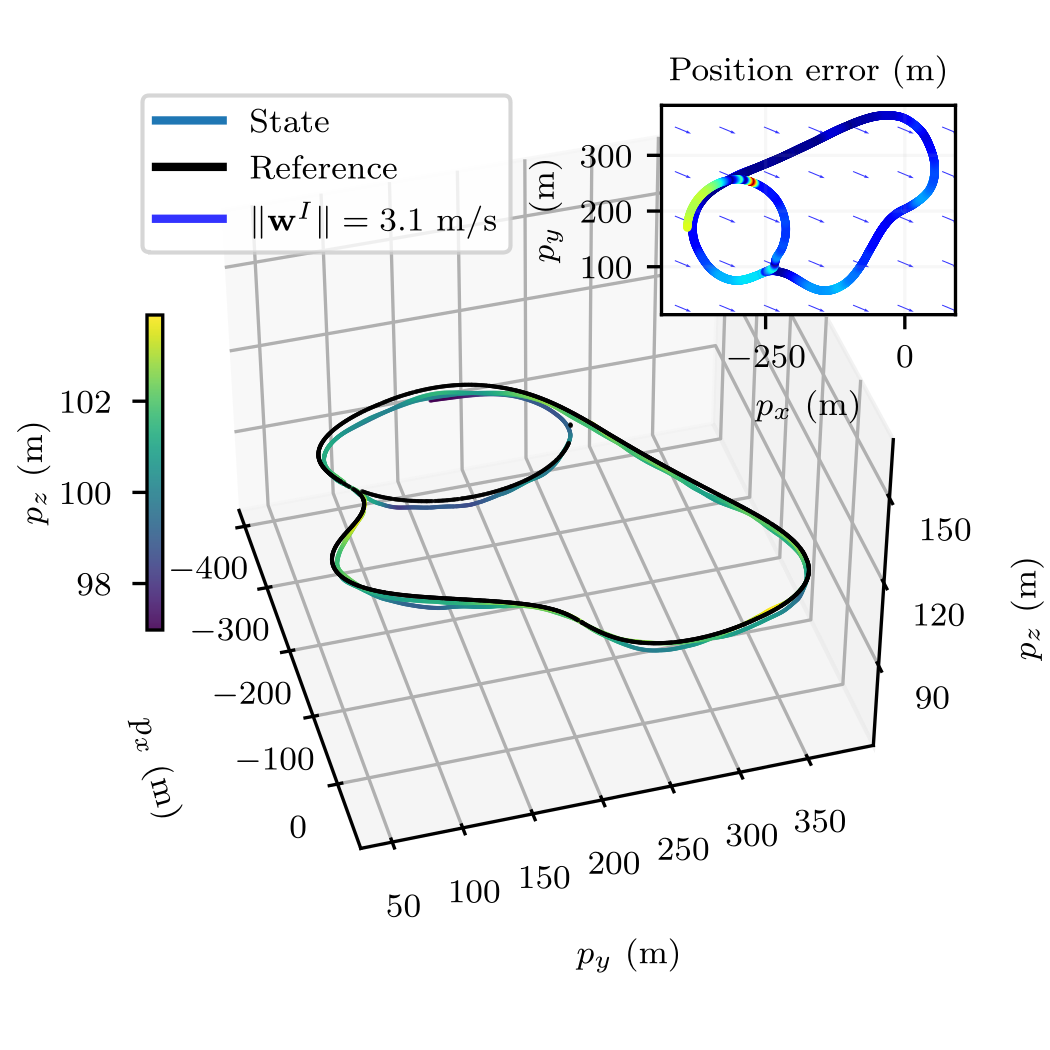}
    
    \subcaption{Position Tracking: prediction vs reference}
  \end{minipage}\hfill
  \begin{minipage}[t]{0.5\textwidth}
    \centering
    \includegraphics[height=7.5cm,width=\linewidth,keepaspectratio,trim=0 0 0 0, clip]{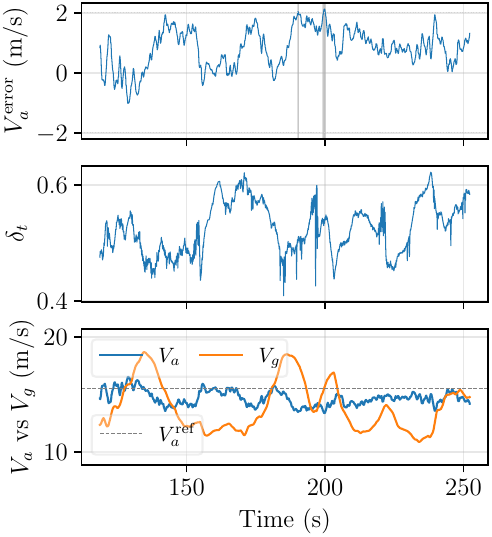}
    \subcaption{Thrust and Velocity Tracking. $V_a^{\text{error}}$ is the absolute error of the airpeed w.r.t. the reference airspeed.}
  \end{minipage}
  \caption{NMPC tracking results (Real-world): (a) position tracking and (b) thrust and velocity tracking with error highlighting. }
  \label{fig:nmpc_results_pos_thrust}
  %\vspace{-10pt}
\end{figure*}

% Table~\ref{tab:nmpc_mean_std} reports the position RMSE (mean and standard deviation) under varying wind magnitudes using the wind-aware sampling strategy, averaged over ten trials. At low wind speeds ($2–6~\si{ms^{-1}}$), the controller achieves error less than $3.7~\si{m}$ with low variance, indicating consistent tracking performance. As wind magnitude increases, the RMSE grows, reflecting the added challenge of countering stronger disturbances. At $8-10~\si{ms^{-1}}$ winds, the highest errors are observed, to be reaching $5.2$~\si{m}. 

We test the robustness of our controller by testing it in not only different wind conditions, but also in known failure modes. In the example shown in Fig.~\ref{fig:Infeasible Rerference}, we inject discontinuity in certain regions in the prescribed motions which makes the reference trajectory physically infeasible for a fixed-wing UAV, violating curvature constraints imposed by the vehicle’s dynamics and actuator limits. This enables us to test the robustness of the controller in predicting feasible trajectories. This is illustrated in the annotated region of Fig.~\ref{fig:Infeasible Rerference}, where tracking the reference is impossible while satisfying the dynamics. In such cases, the dynamics NMPC does not attempt to follow the infeasible command but instead exploits its embedded dynamics constraints to inherently enforce curvature feasibility. This results in a dynamically consistent, curvature-feasible executed trajectory that respects aerodynamic forces, moments, and actuator limits, while staying close to the original path. These results highlight the ability of NMPC not only to maintain robustness in the presence of wind, but also to gracefully handle infeasible references by projecting them into the space of physically realizable solutions.

\subsection{Real-world Validation}
Next, we evaluate the performance of the NMPC formulations in real-world flight experiments. The experiments were carried out to evaluate the robustness of the controller within a test area of $0.15~\text{km}^2$, under wind conditions ranging from $3$ to $10~\si{ms^{-1}}$. Fig.~\ref{fig:nmpc_results_pos_thrust}(a) compares predicted and reference trajectories in the horizontal (x--y) and vertical (z) directions. The top plot shows closed-loop tracking along an elliptical path in the x--y plane, where the color bar indicates the instantaneous position error remains within $3-5$~\si{m} for most of the flight. Overall, the position RMSE remains bounded in both lateral and vertical directions, demonstrating accurate long-horizon tracking in real conditions.  

\textbf{Velocity and thrust regulation.}  
Fig.~\ref{fig:nmpc_results_pos_thrust}(b) shows the coupling between airspeed error, throttle, and wind. Velocity deviations remain small, with errors rising during tailwind phases as $V_g$ increases. The controller compensates by modulating throttle $\delta_t$, increasing under headwind and reducing under tailwind. The shaded grey regions mark instances where the error exceeds $2\si{ms^{-1}}$. In each case, the UAV quickly recovers and stabilizes $V_a$, highlighting effective throttle regulation against wind disturbances. 

% \textbf{Attitude and angular velocity tracking.}  
% Fig.~\ref{fig:nmpc_results_att_angvel} presents attitude-level tracking performance. The left subplot shows the rotation error (see~\cite{huynh2009metrics}), which remains largely around 0.5 rad, with peaks near sharp turns but without divergence. The right subplot shows the RMSE of angular velocity tracking, which stays below $0.4~\si{rads^{-1}}$ and remains well bounded over the $200$~\si{s} horizon. These results confirm that the NMPC maintains stable orientation and angular dynamics, ensuring controllability even under environmental disturbances.

\section{Conclusion}
We presented an NMPC approach for fixed-wing UAV trajectory tracking that combines differential flatness–based trajectory generation with predictive, constraint-aware control. The proposed framework incorporates a dynamic model capturing the aerodynamic forces and moments acting on the vehicle, as well as a novel wind-aware sampling strategy embedded that proactively accounts for wind disturbances to ensure accurate trajectory tracking. Validation in simulation, PX4 SITL, and real-world flight experiments demonstrates that the proposed formulation achieves robust tracking performance even under strong wind disturbances.

Future works will explore the integration of neural dynamics models to further improve system modeling accuracy and the development of energy-aware predictive control strategies to enhance flight efficiency. Finally, given the increased model complexity and broader operating regimes of fixed-wing UAVs compared to multirotors, we will investigate system identification strategies, including data-driven approaches, to enable rapid estimation of platform dynamics for new aircrafts.
\bibliographystyle{IEEEtran}
\bibliography{references}

\end{document}